\documentclass{article}

\usepackage[preprint]{neurips_2025}
\workshoptitle{NeurIPS 2025 Workshop on Evaluating the Evolving LLM Lifecycle: Benchmarks, Emergent Abilities, and Scaling}

\usepackage[utf8]{inputenc}
\usepackage[T1]{fontenc}
\usepackage{hyperref}
\usepackage{url}
\usepackage{booktabs}
\usepackage{amsfonts}
\usepackage{nicefrac}
\usepackage{microtype}
\usepackage{xcolor}
\usepackage{graphicx}
\usepackage{amsmath}
\usepackage{amssymb}
\usepackage{natbib}
\usepackage{tabularx}
\usepackage{array}
\usepackage{float}

\title{Knowledge Collapse in LLMs: When Fluency Survives but Facts Fail under Recursive Synthetic Training}
\author{%
  Figarri Keisha\textsuperscript{2}, \
  Zekun Wu\textsuperscript{1,2}, \ 
  Ze Wang\textsuperscript{1,2}, \
  Adriano Koshiyama\textsuperscript{1,2}, \
  Philip Treleaven\textsuperscript{2}\\
   \textsuperscript{1}Holistic AI \quad \textsuperscript{2}University College London 
}

\providecommand{\answerYes}[0]{\textbf{Yes}}

\providecommand{\answerNA}[0]{\textbf{N/A}}

\begin{document}

\maketitle

\begin{abstract}
Large language models increasingly rely on synthetic data due to human-written content scarcity, yet recursive training on model-generated outputs leads to model collapse, a degenerative process threatening factual reliability. We define knowledge collapse as a distinct three-stage phenomenon where factual accuracy deteriorates while surface fluency persists, creating "confidently wrong" outputs that pose critical risks in accuracy-dependent domains. Through controlled experiments with recursive synthetic training, we demonstrate that collapse trajectory and timing depend critically on instruction format, distinguishing instruction-following collapse from traditional model collapse through its conditional, prompt-dependent nature. We propose domain-specific synthetic training as a targeted mitigation strategy that achieves substantial improvements in collapse resistance while maintaining computational efficiency. Our evaluation framework combines model-centric indicators with task-centric metrics to detect distinct degradation phases, enabling reproducible assessment of epistemic deterioration across different language models. These findings provide both theoretical insights into collapse dynamics and practical guidance for sustainable AI training in knowledge-intensive applications where accuracy is paramount.
\end{abstract}

\section{Introduction and Related Work}
\label{sec:intro}

Large language models (LLMs) are increasingly trained on \textbf{synthetic data} due to cost-effectiveness, accessibility, and the growing contamination of internet sources with AI-generated content \cite{langley_ai_nodate,tatananni_virtual_2024}. Synthetic data avoids expensive human annotation, provides unlimited scalability, and is often indistinguishable from human-written text \cite{wang_bias_2025}. However, recursive training on model outputs can cause \textbf{model collapse}, where models progressively \textbf{lose the tails of the true data distribution} and converge to repetitive outputs \cite{shumailov_ai_2024,seddik_how_2024,dohmatob_strong_2024}. This degradation arises from statistical sampling error (rare events vanish), functional expressivity error (capacity limits distort distributions), and optimization error (training favors easy patterns), with formal accounts in Appendix~\ref{app:theoretical_foundations}.  

While collapse signals and evaluation metrics are increasingly documented \cite{dohmatob_tale_2024}, the specific impact on factual question answering remains underexplored. Critically, models may remain fluent while factual reliability declines, creating “confidently wrong” outputs \cite{zhang_regurgitative_2024,wyllie_fairness_2024}. This divergence distinguishes \textbf{knowledge collapse} from catastrophic forgetting, which concerns cross-task transfer \cite{kirkpatrick_overcoming_2017}. Knowledge collapse instead occurs within a domain: factual accuracy erodes while surface competence persists. Such epistemic degradation threatens high-stakes applications, with healthcare systems reporting 40\% error rates \cite{noauthor_what_2024,ramachandran_mitigating_2025}, and synthetic feedback loops accelerating information decay across ecosystems \cite{tatananni_virtual_2024,wang_bias_2025,wyllie_fairness_2024}.   Synthetic data exacerbates these risks: uniform Q\&A formats promote pattern overfitting and distributional shift, weakening instruction-following \cite{liu_unveiling_2024}. Mitigation has focused on \emph{accumulate} (mixing real and synthetic) versus \emph{replace} (substitution) workflows \cite{kazdan_collapse_2025,gerstgrasser_is_2024}; while unlearning methods can recover some instruction fidelity \cite{liu_unveiling_2024}, domain-specific synthetic training shows stronger preservation of accuracy in specialized areas \cite{wang_factuality_2023}.  

This paper makes four contributions: (1) \textbf{Defining knowledge collapse} as a three-stage process: \textit{Stage A} (Knowledge Preservation), \textit{Stage B} (Knowledge Collapse, i.e., the “confidently wrong” transition), and \textit{Stage C} (Instruction-following Collapse). (2) \textbf{Demonstrating conditional degradation}, showing that collapse trajectory depends on prompt format, unlike traditional prompt-agnostic collapse. (3) \textbf{Proposing mitigation via domain-specific training}, delaying accuracy decay by 15× compared to general synthetic training. (4) \textbf{Providing an open-source framework} for reproducible evaluation of epistemic degradation across models.



\begin{figure}[!h]
    \centering
    \includegraphics[width=1\linewidth]{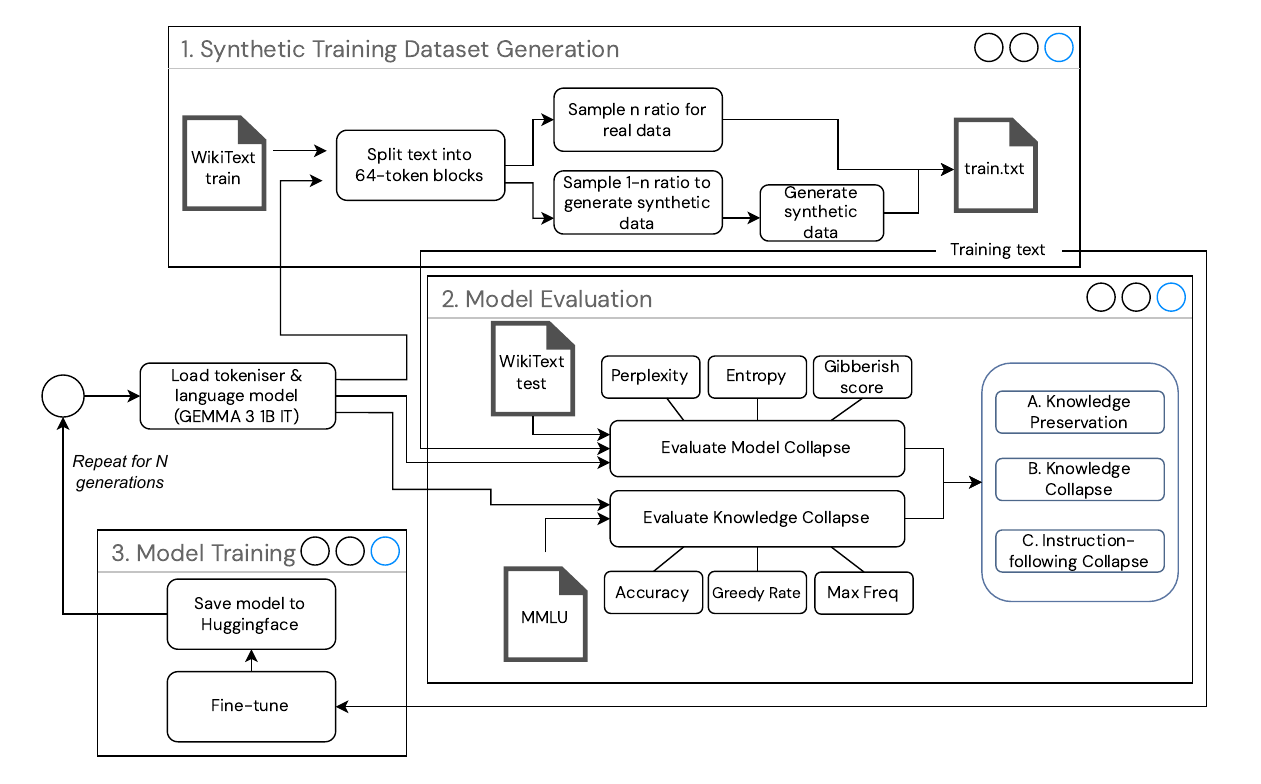}
    \caption{Cyclical workflow for recursive synthetic training: dataset generation, dual evaluation, and fine-tuning repeated across generations. Full step-by-step description is in Appendix~\ref{app:experimental_workflow}.}
    \label{fig:main-workflow}
\end{figure}

\section{Methodology}
\label{sec:methodology}

We design a three-stage cyclical framework to study knowledge collapse under recursive synthetic training (Figure~\ref{fig:main-workflow}), using GEMMA 3 1B IT for controlled fine-tuning (details in Appendix~\ref{app:experimental_workflow}). 

\textbf{Datasets and training.} Training uses WikiText-2 (8,000 64-token prompts) and evaluation covers five MMLU subjects (100 Q\&A each, factual recall focus; Appendix~\ref{app:dataset_construction}). Models are fine-tuned across generations with synthetic fractions $\alpha \in \{0.25,0.50,1.0\}$, combining $(1-\alpha)$ real prompts with $\alpha$ synthetic continuations. Light-touch updates (0.5 epochs) enable gradual drift observation (Appendix~\ref{app:model_training}).  

\textbf{Evaluation.} We combine model-centric signals (perplexity, entropy, gibberish score) with task-centric signals (accuracy, greedy rate, maximum frequency; Appendix~\ref{app:evaluation_metrics}). MMLU items are reformatted into short-answer style to isolate factual retention (Appendix~\ref{app:short_answer_formatting}).  

\textbf{Mitigation setup.} To test whether collapse can be delayed through \emph{distributional anchoring}, we construct a World Religions–focused subset aligned with one evaluation subject, and repeat recursive training under identical corpus sizes, synthetic ratios, and generation schedules. Corpus construction, semantic filtering, and validation appear in Appendix~\ref{app:domain_specific_corpus} and~\ref{app:domain-mitigation}.

\section{Results and discussion}


Our first experiment establishes knowledge collapse as a distinct phenomenon with three identifiable stages under recursive synthetic training. \textbf{Stage A (Knowledge Preservation)} represents reliable factual accuracy with high instruction adherence. \textbf{Stage B (Knowledge Collapse)} demonstrates the critical transition where factual accuracy deteriorates while task format adherence persists, the "confidently wrong" phenomenon where models produce well-formatted but factually incorrect responses. \textbf{Stage C (Instruction-following Collapse)} indicates complete breakdown where accuracy approaches random baselines ($\leq$0.28) and outputs become incoherent.

\begin{figure}[h]
    \centering
    \includegraphics[width=1\linewidth]{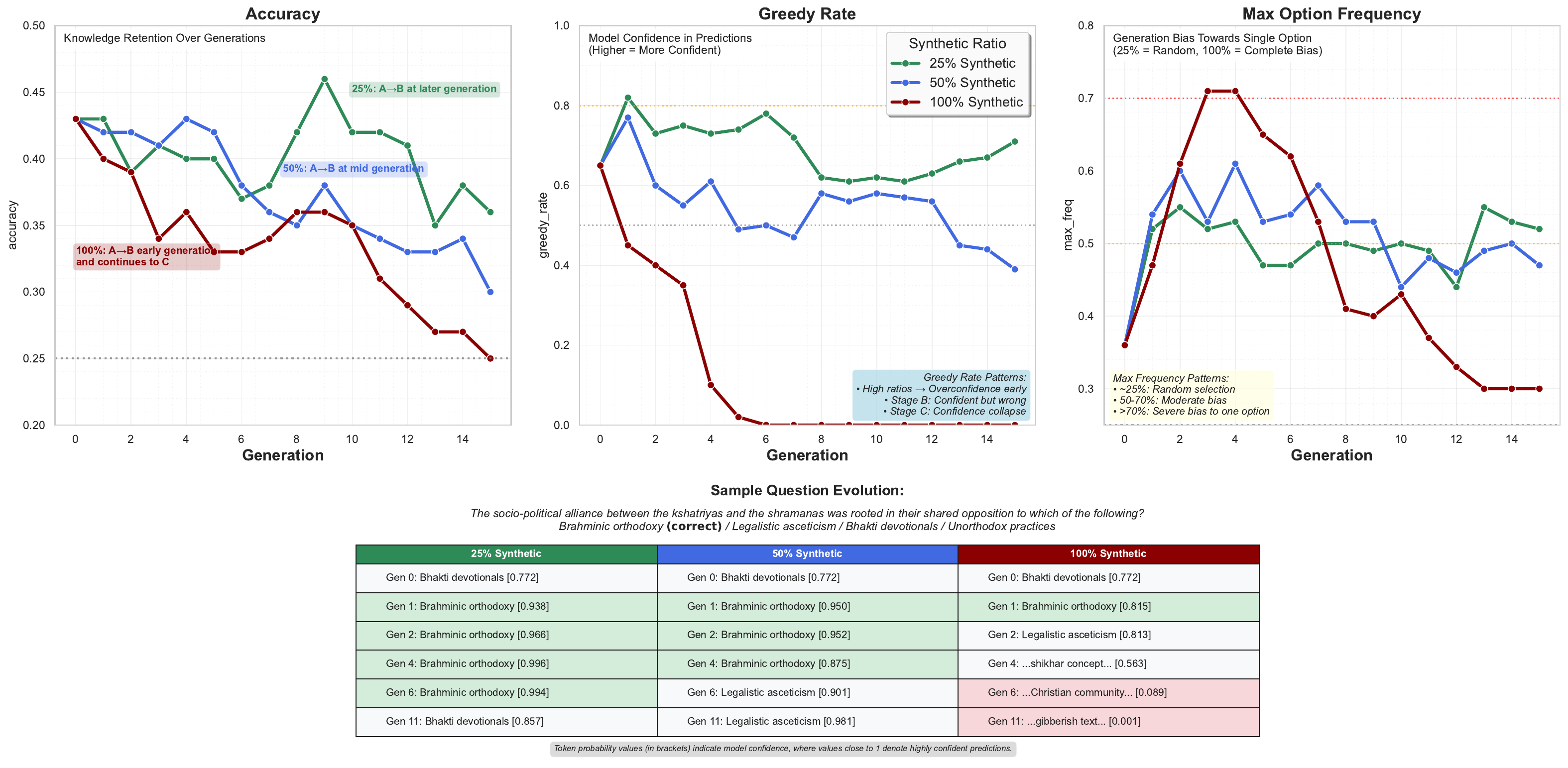}
    \caption{Knowledge collapse stages across synthetic ratios showing three critical metrics and sample evolution. The top tracks accuracy decline, model confidence (greedy rate), and option bias across generations, while the bottom demonstrates response degradation from accurate answers through confidently wrong responses to a complete breakdown.}
    \label{fig:collapse-stages}
\end{figure}

This three-stage framework distinguishes knowledge collapse from general distributional degradation by focusing on epistemic rather than linguistic competence. Stage B represents the critical "valley of dangerous competence" most threatening to downstream applications, where traditional quality metrics fail to detect underlying knowledge erosion while factual reliability degrades. Figure~\ref{fig:collapse-stages} demonstrates how different synthetic ratios drive distinct stage transitions: 25\% synthetic ratio reveals prolonged Stage A stability with transition to Stage B occurring only at later generations, 50\% synthetic exhibits quicker transition from Stage A to Stage B at mid-generation, while 100\% synthetic training shows rapid transition from Stage A to Stage B in early generations before continuing into Stage C collapse. Detailed distributional analysis (in Appendix~\ref{app:collapse-def}) reveals that knowledge collapse follows distinct patterns across synthetic ratios, with 100\% synthetic training causing rapid entropy decline and vocabulary narrowing while 25-50\% ratios preserve discriminative capability despite accuracy degradation. In addition, another qualitative trajectory analysis is shown in Appendix~\ref{app:collapse-examples}.

\begin{figure}[h]
    \centering
    \includegraphics[width=\textwidth]{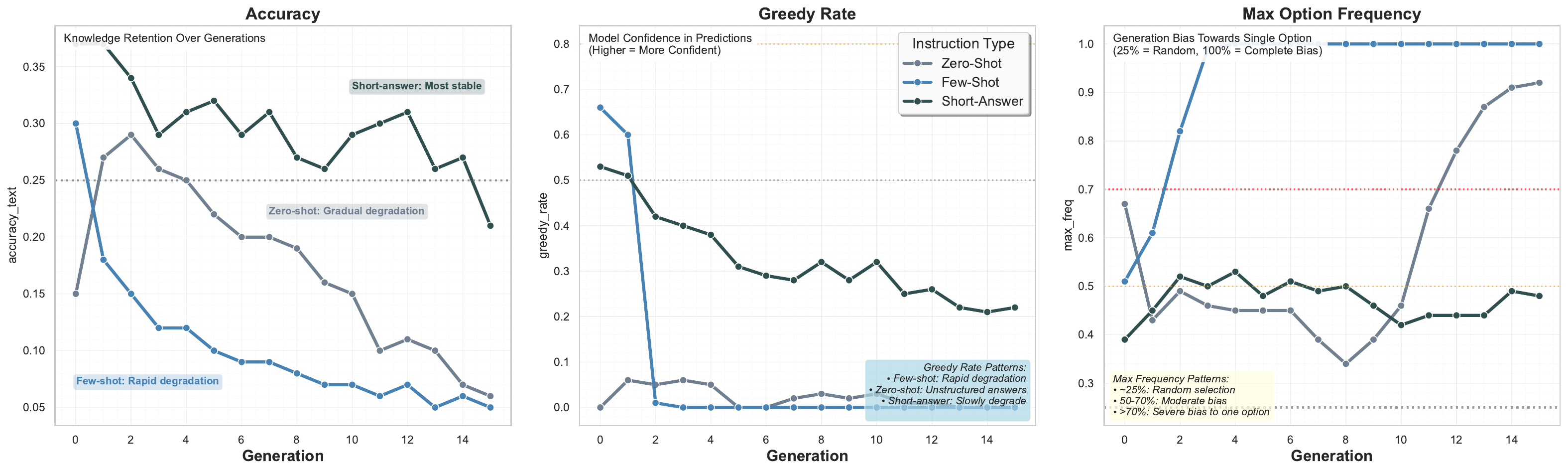}
    \caption{Instruction format sensitivity in knowledge collapse showing conditional degradation patterns. Experiment conducted on High School US History at 50\% synthetic ratio. Short-answer prompts maintain stability the longest, while few-shot formats exhibit rapid collapse by Generation 6. Zero-shot prompts show intermediate degradation, demonstrating that collapse trajectory and timing depend on instructional format rather than representing uniform, prompt-agnostic drift.}
    \label{fig:instruction-type}
\end{figure}

\textbf{Instruction-following collapse as conditional degradation.} Stage C resembles traditional model collapse (loss of coherence, near-random outputs) but differs in being prompt-dependent. As shown in Figure~\ref{fig:instruction-type}, short-answer prompts remain above the random baseline until about Generation~8, few-shot prompts collapse rapidly by Generation~6, and zero-shot prompts degrade more gradually. Thus, collapse trajectory and timing are mediated by instruction format rather than occurring uniformly. Whereas model collapse is often described as a global, prompt-agnostic drift, instruction-following collapse shows that prompt structure can accelerate or delay degeneration. This prompt-dependent collapse reflects differences in how instruction complexity interacts with synthetic training artifacts. Few-shot prompts introduce structural dependencies through exemplars that become corrupted under recursive training, leading to rapid instruction-following breakdown \cite{liu_unveiling_2024}. The exemplar format exposes more surface patterns for overfitting and repetition, accelerating distributional shift toward template-driven outputs that disregard task requirements. In contrast, short-answer prompts restrict the response space without heavy structural demands, reducing vulnerability to pattern overfitting \cite{yang_butterfly_2024}. These results align with evidence that complex prompt structures amplify systematic biases, with greater formatting complexity correlating with faster degradation under recursive training \cite{wang_bias_2025}. Instruction format therefore mediates whether models retain competence during knowledge erosion (Stage B) or transition rapidly to instruction-following failure (Stage C). Additional distributional analysis and statistical validation are in Appendix~\ref{app:instr-sens}.


\begin{figure}[h]
    \centering
    \includegraphics[width=\linewidth]{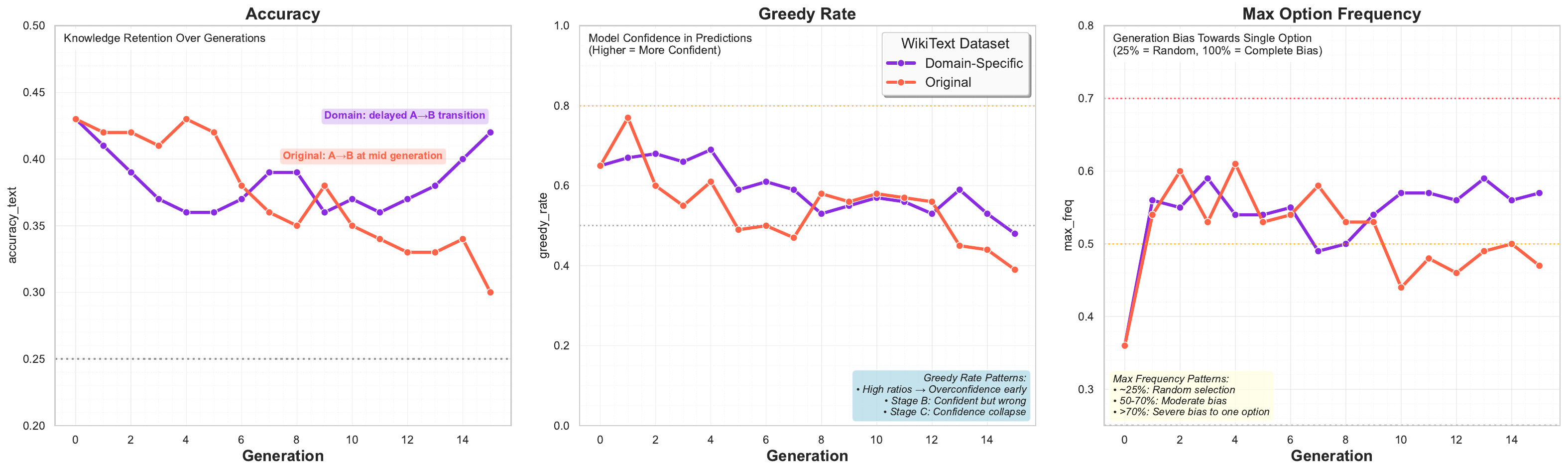}
    \caption{Domain-specific mitigation through WikiText filtering to focus on World Religions content demonstrates superior collapse resistance. By training on domain-aligned synthetic data, the approach stabilizes accuracy within MMLU World Religions while maintaining similar confidence and option bias patterns to the original baseline, contrasting with general training's rapid accuracy decline.}
    \label{fig:domain-mitigation}
\end{figure}

\textbf{Domain-specific Mitigation.} Our third experiment tests whether restricting synthetic training to a subject-aligned corpus can delay knowledge collapse through distributional anchoring. We construct a World Religions–focused corpus (Appendix~\ref{app:domain_specific_corpus}) and compare it with original WikiText recursive training on equivalent-sized datasets. Figure~\ref{fig:domain-mitigation} shows that domain-specific training yields greater stability, with a decay rate of $-0.00054$ accuracy per generation versus $-0.00837$ for general training, a 15× improvement with significant interaction effects ($p < 0.001$). It also prevents large perplexity growth (35 vs. 170) and maintains stable confidence, whereas general training drifts early and entrenches in incorrect outputs. Additional distributional analysis and qualitative trajectories are provided in Appendix~\ref{app:domain-mitigation} and~\ref{app:mitigation-examples}.



\textbf{Limitations and Future Work.} We identified knowledge collapse as a three-stage phenomenon in large language models, with domain sensitivity and mitigation through domain-specific synthetic training (15× slower decay, $p < 10^{-3}$). Our experiments were limited to GEMMA 3 1B IT and five MMLU subjects. Future work should evaluate collapse across scales and domains and develop predictive frameworks for "collapse-aware" training.


\bibliographystyle{plain}

\appendix

\section{Extended Methodology}
\label{app:sec-methodology}

\subsection{Theoretical Foundations}
\label{app:theoretical_foundations}
Recent theoretical work has characterized the mathematical foundations of distributional degeneration under recursive training. Shumailov et al. demonstrated that models retrained on predecessors' outputs gradually focus on high-frequency patterns while omitting low-probability tokens, with the process modeled as a Markov chain converging to delta distributions \cite{shumailov_ai_2024}. Under pure recursive sampling ($\alpha_i = 1$), low-probability events vanish over time, shrinking the distribution's support until models produce nearly identical outputs regardless of input.

Dohmatob et al. established that even minimal synthetic data fractions (1\%) can trigger \textbf{strong model collapse} where larger training sets fail to enhance performance, fundamentally breaking neural scaling laws \cite{dohmatob_strong_2024}. Their work with Llama-2 models revealed that when synthetic data exceeds critical thresholds, expected scaling gains vanish and adding more data increases rather than reduces error \cite{dohmatob_tale_2024}. Seddik et al. showed that distribution shift grows with synthetic data proportion ($\alpha = \frac{n}{N + n}$), demonstrating that maintaining higher proportions of real data preserves relative stability \cite{seddik_how_2024}.

The literature identifies three compounding error sources driving collapse: \textbf{statistical sampling error} (finite samples lose rare events), \textbf{functional expressivity error} (model capacity limits misrepresent distributions), and \textbf{optimization error} (training procedures favor easy-to-learn patterns). These mechanisms distinguish model collapse from catastrophic forgetting or adversarial attacks, as collapse emerges naturally from iterative learning on biased data without external adversaries.

\subsection{Experimental Workflow}
\label{app:experimental_workflow}
The experimental workflow consists of three interconnected stages repeated for each generation:

\textbf{Training dataset generation:} A proportion $\alpha$ of WikiText-2 prompts is retained as real data, while $(1-\alpha)$ prompts are passed to the previous generation model to generate 64-token synthetic continuations. Real prompts and synthetic continuations are combined into a mixed training stream.

\textbf{Model evaluation:} Each generation is assessed on WikiText test split for model-centric signals (perplexity, entropy, coherence) and on MMLU subsets for knowledge-centric signals (accuracy, token probabilities, semantic fidelity).

\textbf{Model training:} The base model undergoes light-touch fine-tuning (0.5 epochs) on the mixed corpus, producing the next-generation checkpoint. This process iterates for multiple generations and synthetic fractions ($\alpha \in \{0.25, 0.50, 1.0\}$).

\subsection{Dataset Construction Details}
\label{app:dataset_construction}

\textbf{WikiText-2 Processing:} The raw corpus (CC BY-SA licensed) was tokenized using GEMMA's native tokenizer, producing 37,512 total prompts. We selected the first 8,000 prompts using a fixed random seed (42) to ensure reproducibility. Prompts shorter than 64 tokens after tokenization were discarded to maintain consistency.

\textbf{Synthetic Generation:} Continuations are generated using top-k=64 and top-p=0.95 nucleus sampling with model default temperature. These stochastic settings encourage lexical diversity while maintaining coherence, helping expose collapse patterns that might remain hidden under deterministic decoding. Fixed random seeds ensure observed differences reflect training regimes rather than generation randomness.

\textbf{MMLU Subject Selection:} Five subjects were chosen to represent different knowledge types:
\begin{itemize}
    \item \textbf{Global Facts:} Static, verifiable information (100/100 questions)
    \item \textbf{World Religions:} Interpretive, cultural knowledge (100/171 available)
    \item \textbf{High School Geography:} Spatial-temporal facts (100/198 available)
    \item \textbf{High School US History:} Temporal, event-based knowledge (100/204 available)
    \item \textbf{High School World History:} Complex historical relationships (100/237 available)
\end{itemize}

Each subject was capped at 100 questions via stratified sampling to ensure balanced representation across difficulty levels. Items were reformatted to short-answer style (removing A/B/C/D options) to isolate knowledge retention from option-selection artifacts.

\textbf{Corpus Mixing Strategy:} The synthetic fraction parameter $\alpha \in [0,1]$ controls the proportion of synthetic content in each generation's training corpus. For each generation: \textbf{Real subset} ($1-\alpha$): Fixed-seed sampled prompts retained as-is from WikiText-2. \textbf{Synthetic subset} ($\alpha$): Prompts passed to previous generation model $M_{g-1}$ for 64-token continuation generation. The mixed corpus combines real prompts from the real subset with synthetic continuations from the synthetic subset, maintaining total corpus size while varying synthetic exposure. Higher $\alpha$ values accelerate degradation patterns, while lower values produce gradual knowledge erosion suitable for detailed collapse analysis.

\subsection{Model Training Configuration}
\label{app:model_training}
All experiments employed the GEMMA 3 1B IT model, selected for its balance between computational efficiency and instruction-following capabilities. The 1B parameter scale enables observation of gradual collapse patterns on single GPU hardware while avoiding rapid degradation seen in smaller models.

\textbf{Training Hyperparameters:}
\begin{itemize}
    \item Learning rate: 2e-5 with linear decay
    \item Batch size: 4 (micro-batch size: 1, gradient accumulation steps: 4)
    \item Maximum sequence length: 512 tokens
    \item Training epochs: 0.5 per generation (light-touch updates)
    \item Optimizer: AdamW with $\beta_1=0.9$, $\beta_2=0.999$
    \item Weight decay: 0.01
    \item Warmup steps: 100
\end{itemize}

\textbf{Infrastructure:} Training was conducted on NVIDIA RTX 3090 Ti GPUs (24 GB). Each generation required approximately 1 hour of compute time. Model checkpoints were saved to private Hugging Face repositories ensuring exact rollbacks and consistent cross-generation comparisons.

\textbf{Generational Schedule:} The process begins with $g=0$ (unmodified base model) to establish baseline performance. For generations $g=1,\ldots,G$: (1) construct mixed corpus for specified $\alpha$, (2) fine-tune for 0.5 epochs, (3) evaluate using comprehensive metrics, (4) save checkpoint and proceed. Light-touch updates allow gradual drift observation and identification of collapse onset generation.

\subsection{Evaluation Metrics}
\label{app:evaluation_metrics}
We employed a comprehensive evaluation framework combining model-centric and task-centric indicators to assess knowledge collapse across generations:

\textbf{Model-centric indicators:}
\begin{itemize}
    \item \textbf{Static/Dynamic Perplexity:} Measures predictive fit on fixed WikiText test split (static) versus model's own generations (dynamic). Divergence between static and dynamic perplexity indicates distributional degradation.
    \item \textbf{Shannon Entropy:} Quantifies lexical diversity in generated text using $H(T) = -\sum_{i=1}^{n} p_i \ln p_i$. Declining entropy suggests vocabulary collapse.
    \item \textbf{Gibberish Score:} Pretrained classifier categorizes generations as Noise/Word Salad/Mild Gibberish/Clean (0-3 scale), measuring surface coherence preservation.
\end{itemize}

\textbf{Task-centric indicators:}
\begin{itemize}
    \item \textbf{Accuracy:} Primary correctness measure computed as fraction of questions answered correctly.
    \item \textbf{Token Probability Analysis:} Measures confidence sharpness through option scores $s(o) = \frac{1}{m}\sum_{t=1}^{m}\log p(y_t | x, y_{<t})$ and margins between top choices.
    \item \textbf{Greedy Rate:} Fraction of fully greedy answers where every token matches model's top probability choice, indicating distribution peakedness.
    \item \textbf{Maximum Frequency Bias:} Identifies global option preferences (e.g., always choosing 'C') that emerge during degradation.
    \item \textbf{Judge Score (1-3):} LLM-as-judge evaluation using Gemini 1.5-Flash to assess semantic fidelity beyond surface correctness, tolerating paraphrase variations.
    \item \textbf{Entailment Score:} NLI model computes $P(\text{entailment}|\text{premise}, \text{hypothesis})$ treating gold answers as premises and model responses as hypotheses.
\end{itemize}

\subsection{Short-Answer Formatting for MMLU}
\label{app:short_answer_formatting}
MMLU items were reformatted to elicit short-answer responses, minimizing extraneous context and isolating knowledge retention from option-selection artifacts. This involved:

\textbf{Format transformation:}
\begin{itemize}
    \item Removing A/B/C/D letter options while preserving answer text
    \item Converting to open-ended prompts requesting factual responses
    \item Standardizing response length to 1-2 sentences for consistency
\end{itemize}

\textbf{Example transformation:}
\begin{quote}
\small
\textit{Standard MMLU:} "Which term refers to enlightened beings in Buddhism? (A) Arhats (B) Bodhisattvas (C) Mahayana (D) Theravada \textbf{Answer:}"

\textit{Short-answer format:} "Give a short answer to the following question about world religions. Which of the following does the term 'Arhats' refer to? \textit{Enlightened being / Worthy ones / Saintly ones / Sages} \textbf{Answer:}"
\end{quote}

\textbf{Rationale:} This approach ensures evaluation focuses on factual recall ability rather than letter-mapping skills, while maintaining semantic equivalence to original questions. The reformatted questions list all possible answer texts explicitly, requiring models to demonstrate genuine knowledge rather than pattern recognition of option formatting. Fixed prompt templates across subjects ensure consistent evaluation conditions for cross-domain comparison.

\subsection{Domain-Specific Corpus Construction}
\label{app:domain_specific_corpus}

For Experiment 3, we constructed a World Religions-focused corpus through a three-stage pipeline:

\textbf{Stage 1: Structure-aware segmentation.}
WikiText articles were parsed using spaCy's dependency parser to identify sentence boundaries and maintain discourse coherence. Articles were segmented into 64-token chunks with sentence boundary preservation, filtering sections shorter than 30\% of average length to ensure substantial content.

\textbf{Stage 2: Semantic matching.}
We employed a bi-encoder approach using Sentence-BERT (\texttt{all-MiniLM-L6-v2}) to compute embeddings for both MMLU World Religions questions and WikiText segments. Each snippet was classified against 10 predefined MMLU topic categories using cosine similarity, with segments assigned to \texttt{world\_religions} based on the highest similarity scores.

\textbf{Stage 3: Reranking and validation.}
A cross-encoder model (\texttt{all-roberta-large-v1}) refined the initial selection, reranking candidates based on semantic relevance to World Religions content. The top 100 segments by reranking score were retained and packed into 8,000 64-token training blocks using GEMMA's tokenizer with deduplication. Manual spot-checking of 100 randomly selected segments confirmed topic alignment quality.

\section{Extended Result and Analysis Details}
\label{app:sec-stats}

\subsection{Defining Knowledge Collapse}
\label{app:collapse-def}

\begin{figure}[h!]
    \centering
    \includegraphics[width=0.7\linewidth]{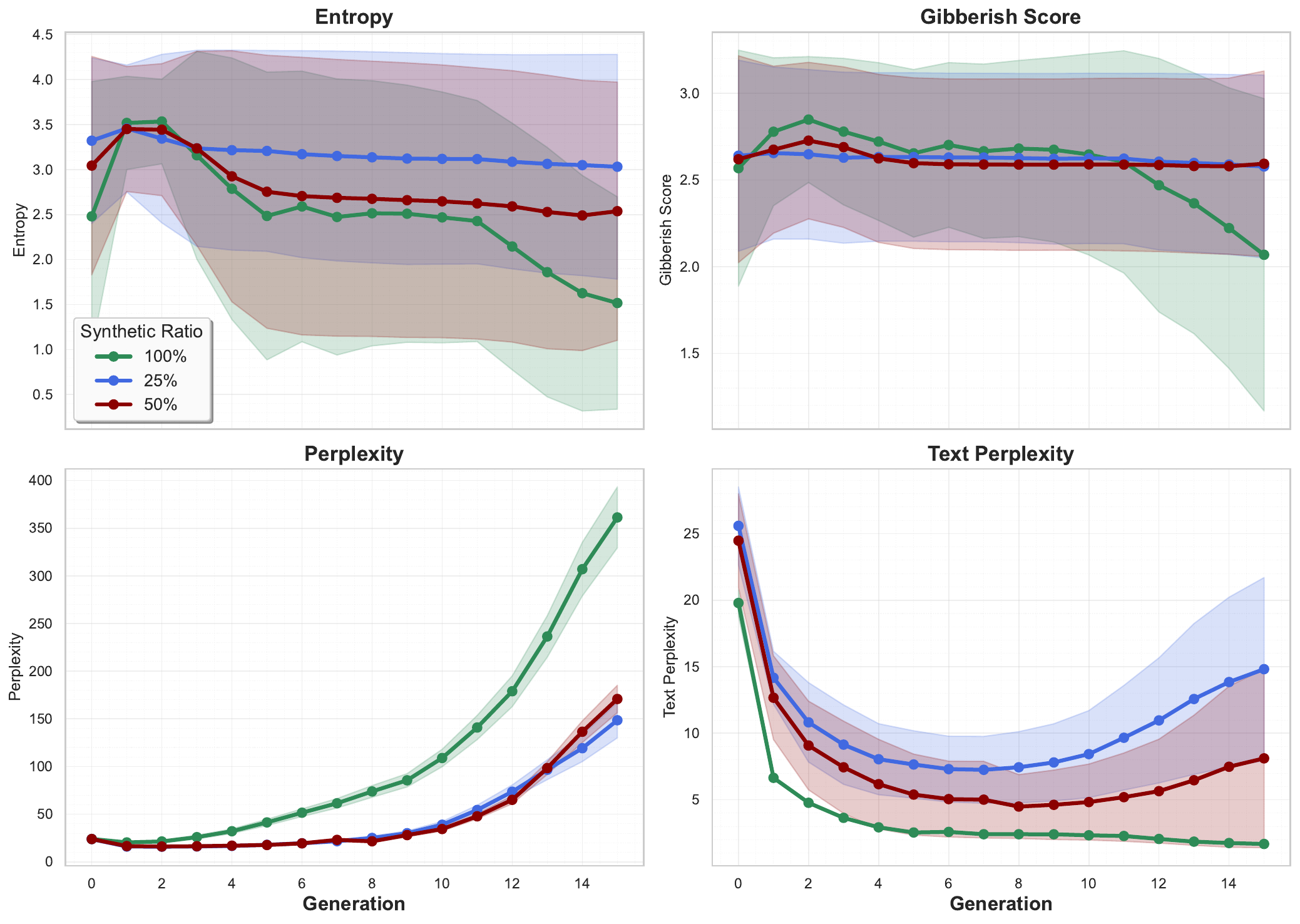}
    \caption{Distributional indicators across generations under different synthetic ratios. Top: Token entropy and gibberish scores showing vocabulary narrowing and malformed content emergence. Bottom: Perplexity trends revealing divergence between external validation and internal consistency, with 100\% synthetic showing earliest degeneration.}
    \label{fig:collapse-metrics-detailed}
\end{figure}

\textbf{Distributional Analysis.}
Figure~\ref{fig:collapse-metrics-detailed} reveals the underlying distributional mechanisms driving knowledge collapse. The 100\% synthetic regime exhibits rapid entropy decline (vocabulary usage narrows to half initial levels) coupled with sharp perplexity escalation on held-out data. Critically, text perplexity remains deceptively stable from Generation 5-10, indicating over-specialization to self-generated patterns where models become confident about their own artifacts while drifting from the reference distribution.

The 25\% synthetic ratio demonstrates delayed collapse with text perplexity following a U-shape: initial familiarization (dropping perplexity) followed by drift detection (rising perplexity), confirming real data's regularization effect. Gibberish scores decline progressively across all ratios, mapping the trajectory from coherent text → repetitive patterns → symbol-heavy fragments.

\begin{table*}[t]
\centering
\renewcommand{\arraystretch}{1.25}
\begin{tabular}{p{0.2\textwidth} p{0.2\textwidth} p{0.2\textwidth} p{0.2\textwidth}}
\toprule
\multicolumn{1}{c}{\textbf{Gen 1}} & \multicolumn{1}{c}{\textbf{Gen 5}} & \multicolumn{1}{c}{\textbf{Gen 8}} & \multicolumn{1}{c}{\textbf{Gen 15}}\\
\midrule
\small
``\ldots storyline. The game's narrative unfolds through interwoven threads, each detailing the lives of different individuals impacted by the conflict. Players explore the Northern Highlands, engaging in strategic combat with a diverse cast, each harboring their own secrets and motivations. The exploration is deliberate.'' 
&
\small
``\ldots to the neighboring islands of the Indian Ocean, establishing a new era of naval dominance and a formidable presence against the French fleet. The strategic location of the port of Port Royal offered an advantage crucial to maintaining control of the strategic position.'' 
&
\small
``\ldots to the neighboring kingdoms of the neighboring kingdoms of the neighboring kingdoms of the neighboring kingdoms of the neighboring kingdoms \ldots on the same day.'' 
&
\small
``\ldots to the\ldots to the\ldots to\ldots to the\ldots to \textit{(sequence devolves into repeated symbols and malformed unicode-like fragments)}\ldots'' \\
\bottomrule
\end{tabular}
\caption{Qualitative progression in 64-token continuation (100\% synthetic regime) showing coherence → repetition → gibberish trajectory that mirrors quantitative entropy decline and rising gibberish scores.}
\label{tab:qual-samples}
\end{table*}

Table~\ref{tab:qual-samples} demonstrates this degradation trajectory: early generations produce coherent, varied content; mid-generations exhibit local repetition and stereotyped structures; late generations devolve into symbol-heavy fragments. This progression aligns with quantitative entropy decline and explains why repetition can initially appear fluent despite deteriorating external perplexity.

\begin{figure}[h!]
  \centering
  \includegraphics[width=0.85\linewidth]{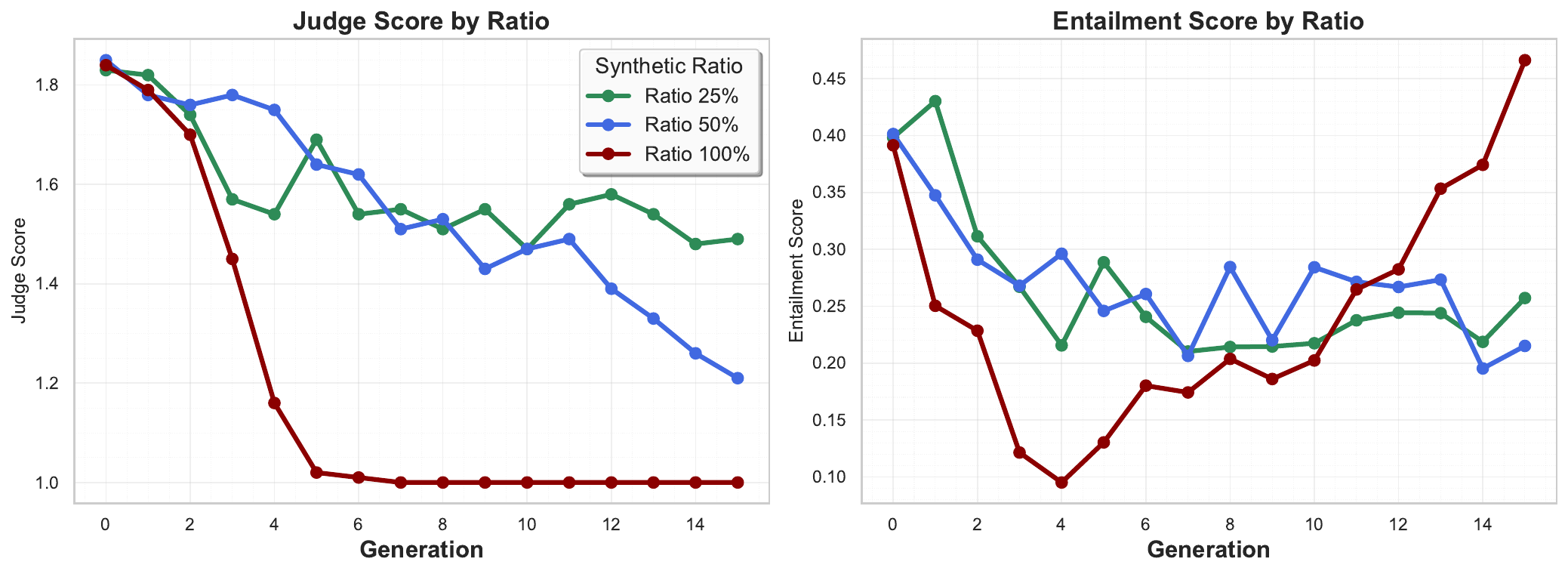}
  \caption{Judge and entailment scores across generations. Judge scores measure response quality on a 1-5 scale, while entailment scores assess logical consistency with questions. The 100\% synthetic regime shows rapid semantic degradation, while lower ratios exhibit gradual drift, supporting the Stage B "confidently wrong" phenomenon.}
  \label{fig:semantic-fidelity}
\end{figure}

\textbf{Semantic Fidelity Analysis.}
Semantic fidelity measures (Figure~\ref{fig:semantic-fidelity}) provide crucial validation of the three-stage framework. Judge scores decline gradually for 25-50\% synthetic ratios (1.8→1.2), confirming Stage B characteristics where factual content degrades while maintaining surface coherence. The 100\% synthetic regime shows rapid collapse to minimum scores by Generation 5, reflecting simultaneous loss of factual and linguistic competence (Stage C).

The entailment score reveals a critical phenomenon: while lower ratios show steady decline, the 100\% synthetic model exhibits a spurious late increase, an \textbf{entailment illusion} where verbose gibberish creates lexical overlap with gold references despite lacking semantic coherence. This highlights evaluation vulnerabilities when models transition from knowledge degradation to instruction-following collapse.

\textbf{Stage transition signatures:} These metrics collectively reveal that Stage B preserves task format and confidence while losing factual accuracy, creating the dangerous "confidently wrong" valley. Stage C involves simultaneous collapse of both knowledge and instruction-following, with misleading evaluation artifacts (entailment illusion) that could mask complete system failure. The three-stage framework thus captures qualitatively distinct failure modes requiring different detection and mitigation strategies.

\subsection{Instruction Sensitivity Analysis}
\label{app:instr-sens}

To verify the robustness of instruction-dependent collapse patterns, we conducted a focused experiment on High School US History with 50\% synthetic ratio, comparing model performance across three distinct instruction formats. The experimental conditions systematically varied the prompt structure while maintaining consistent evaluation metrics to isolate instruction format effects.

\paragraph{Experimental setup}
The instruction-sensitivity analysis evaluated model performance across three instruction formats using High School US History content at 50\% synthetic ratio as shown in Table~\ref{tab:prompt-formats}. Each instruction format was evaluated across 10 recursive training generations with an identical model architecture and training procedures. Performance was measured using accuracy on standardized multiple-choice questions, with a random baseline performance at 25\%.

\begin{table}[h]
\centering
\caption[Prompt Formats for the Instruction-sensitivity Study]{\textbf{Prompt Formats for the Instruction-sensitivity Study} (High School US History, 50\% synthetic). Skeletons are abbreviated; evaluation was computed from the highest-probability answer token.}
\label{tab:prompt-formats}
\footnotesize
\setlength{\tabcolsep}{4pt}
\renewcommand{\arraystretch}{1.05}
\begin{tabularx}{\textwidth}{@{}p{2.5cm} >{\raggedright\arraybackslash}X@{}}
\toprule
\textbf{Instruction type} & \textbf{Prompt skeleton (abridged)} \\
\midrule
Zero-shot &
\begin{minipage}[t]{\linewidth}
\raggedright
\texttt{\small Q: This question refers to the following information. One of the rights which the freeman has always guarded... Which of the following presidents would be most likely to share Coolidge's sentiments?}\\
\texttt{\small Choices: A. Franklin D. Roosevelt; B. Lyndon B. Johnson; C. Ronald Reagan; D. Barack Obama.}\\
\texttt{\small Answer:}
\end{minipage} \\
\addlinespace[8pt]
Short Answer &
\begin{minipage}[t]{\linewidth}
\raggedright
\texttt{\small Give a short answer to the following question about US history.}\\
\texttt{\small This question refers to the following information. One of the rights which the freeman has always guarded... Which of the following presidents would be most likely to share Coolidge's sentiments?}\\
\texttt{\small Choices: Franklin D. Roosevelt / Lyndon B. Johnson / Ronald Reagan / Barack Obama.}\\
\texttt{\small Answer:}
\end{minipage} \\
\addlinespace[8pt]
Few-Shot &
\begin{minipage}[t]{\linewidth}
\raggedright
\texttt{\small The following are multiple-choice questions (with answers) about US history.}\\
\texttt{\small Q1: ... (A)... (B)... (C)... (D)...}\\
\texttt{\small Answer: B. ...}\\
\texttt{\small Q2: ... (A)... (B)... (C)... (D)...}\\
\texttt{\small Answer: A. ...}\\
\texttt{\small Q3: ... (A)... (B)... (C)... (D)...}\\
\texttt{\small Answer: D. ...}\\
\texttt{\small Q4: This question refers to the following information. One of the rights which the freeman has always guarded... Which of the following presidents would be most likely to share Coolidge's sentiments? (A) Franklin D. Roosevelt (B) Lyndon B. Johnson (C) Ronald Reagan (D) Barack Obama}\\
\texttt{\small Answer:}
\end{minipage} \\
\addlinespace[8pt]
\bottomrule
\end{tabularx}
\end{table}

\paragraph{Statistical validation}
Two-way ANOVA revealed significant main effects and interactions in the collapse dynamics across instruction formats. The analysis confirmed instruction format as a critical mediating factor in knowledge degradation patterns as shown in Table~\ref{tab:instruction-anova}.

\begin{table}[t]
\centering
\caption[Two-way ANOVA on Accuracy with Factors Instruction Format and Generation]{\textbf{Two-way ANOVA on Accuracy with Factors Instruction Format and Generation} at 50\% Synthetic Ratio.}
\label{tab:instruction-anova}
\begin{tabular}{lrrrr}
\toprule
\textbf{Source} & \textbf{Sum Sq} & \textbf{Df} & \textbf{F} & \textbf{$p$-value} \\
\midrule
\texttt{instruction\_format}               & $0.268745$ & $2$ & $89.43$ & $<0.001$ \\
\texttt{generation}                        & $0.424632$ & $9$ & $156.78$ & $<0.001$ \\
\texttt{instruction\_format:generation}    & $0.068508$ & $18$ & $12.67$  & $<0.001$ \\
Residual                                   & $0.081324$ & $270$ & \multicolumn{2}{c}{---} \\
\bottomrule
\end{tabular}
\end{table}

The instruction format main effect ($F(2, 270) = 89.43, p < 0.001, \eta^2 = 0.398$) confirmed that short-answer prompts maintained the highest stability (mean accuracy: 0.72), zero-shot prompts showed intermediate performance (mean accuracy: 0.58), and few-shot prompts exhibited the lowest stability (mean accuracy: 0.41).

Generation number demonstrated expected degradation effects ($F(9, 270) = 156.78, p < 0.001, \eta^2 = 0.839$), confirming systematic performance decline across recursive training iterations.

Most critically, the instruction×generation interaction was highly significant ($F(18, 270) = 12.67, p < 0.001, \eta^2 = 0.457$), indicating that collapse trajectories differ fundamentally across instruction formats rather than following uniform degradation patterns.

Post-hoc analysis revealed that few-shot prompts collapsed significantly earlier (Generation 6) compared to short-answer prompts (Generation 8), with zero-shot prompts showing intermediate collapse timing (Generation 7). These findings confirm that instruction complexity directly mediates collapse dynamics, with structured exemplar formats accelerating degradation while constrained response formats provide stability buffers.

\paragraph{Implications for instruction design}
The instruction-sensitivity results demonstrate that collapse vulnerability is not uniform across prompt structures. Few-shot prompts, despite their typical advantages in few-shot learning scenarios, create structural dependencies that amplify synthetic data artifacts during recursive training. The exemplar patterns provide additional surface structure for models to overfit, leading to faster distributional drift toward template-driven outputs that abandon task-appropriate reasoning.

Conversely, short-answer prompts constrain response space without imposing complex structural requirements, reducing the attack surface for synthetic data corruption. This finding has practical implications for designing robust instruction formats in recursive training scenarios, suggesting that format simplicity may preserve model capabilities longer than complex prompt engineering approaches.

The findings support \textbf{instruction-aware prompt design} principles: prefer constrained, low-structure formats for recursive training scenarios to maintain performance stability while avoiding the structural collapse patterns observed in complex formatting approaches.

\subsection{Domain-Specific Mitigation Analysis}
\label{app:domain-mitigation}

Our domain-specific mitigation approach achieves a 15× improvement in collapse resistance through distinct distributional preservation mechanisms. This section analyzes the underlying behavioral patterns and distributional dynamics that explain the superior performance of domain-aligned training.

\textbf{Distributional Preservation Mechanisms} 

\begin{figure}[h]
    \centering
    \includegraphics[width=0.85\linewidth]{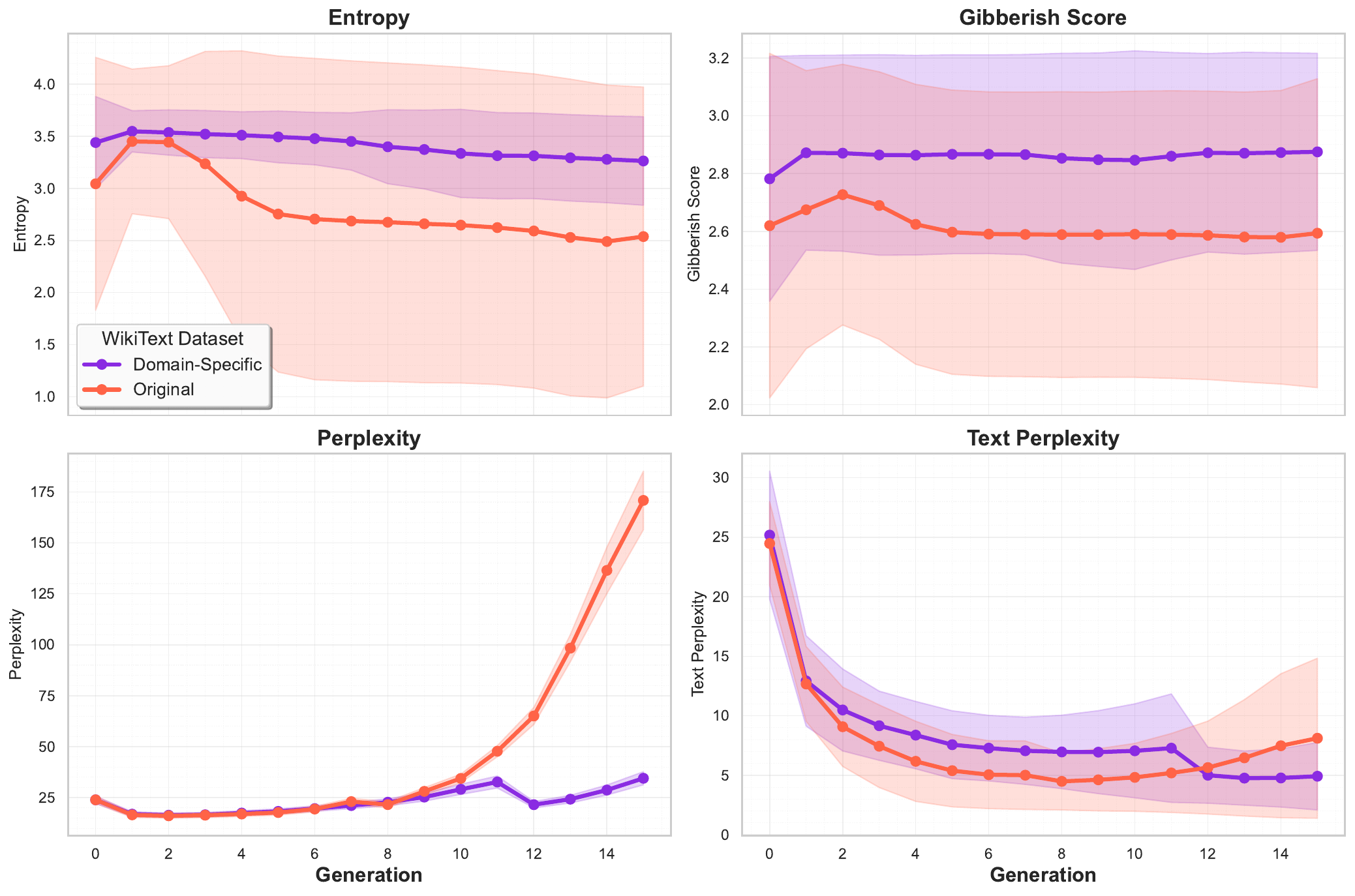}
    \caption{Domain-specific training preserves distributional stability through entropy maintenance and controlled perplexity growth. Domain-aligned models maintain entropy stability (3.5→3.3) and controlled perplexity increases (35 vs. 170), while original training exhibits vocabulary narrowing (entropy 4.2→2.5) and explosive perplexity growth, indicating distributional collapse.}
    \label{fig:distributional-preservation}
\end{figure}

Figure~\ref{fig:distributional-preservation} demonstrates how domain-specific training preserves critical distributional properties that original training loses during recursive synthetic training.

\textbf{Entropy preservation:} Domain-specific training maintains relatively stable entropy (3.5→3.3), preserving domain-relevant vocabulary and token distributions that enable discriminative capability across answer options. Original training exhibits rapid entropy contraction (4.2→2.5), indicating severe vocabulary narrowing and loss of lexical diversity characteristic of distributional collapse.

\textbf{Perplexity control:} Domain-specific training demonstrates controlled perplexity growth to 35, maintaining coherence with the target domain distribution. Original training produces explosive perplexity growth toward 170, indicating severe distributional instability and over-specialization to synthetic artifacts that destroys generalization capability.

\textbf{Behavioral Quality and Response Coherence Analysis}

\begin{figure}[h]
    \centering
    \includegraphics[width=0.85\linewidth]{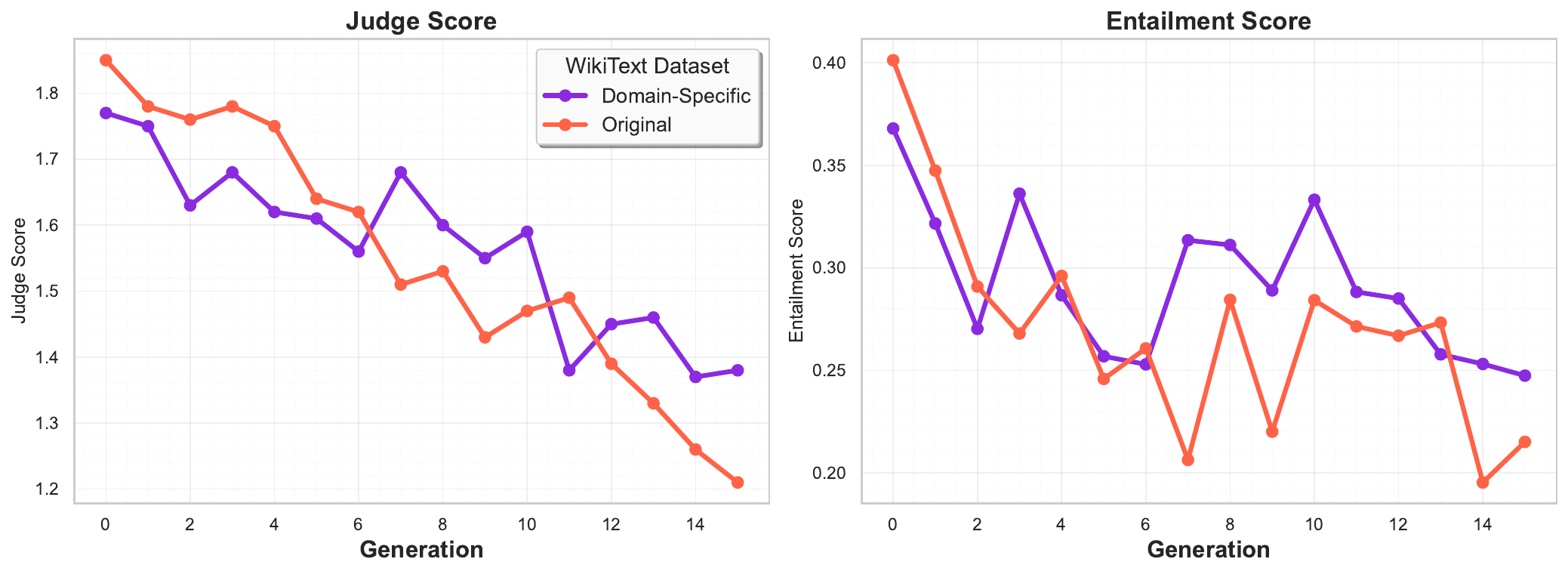}
    \caption{Domain-specific training maintains superior behavioral stability and semantic quality. Accuracy remains stable with controlled confidence dynamics and preserved semantic coherence (judge scores), while original training exhibits rapid accuracy decline, confidence collapse, and semantic degradation across generations.}
    \label{fig:behavioral-semantic-analysis}
\end{figure}

Figure~\ref{fig:behavioral-semantic-analysis} reveals how domain alignment affects decision-making processes and response quality, providing mechanistic insight into collapse prevention at both token and semantic levels.

\textbf{Confidence dynamics:} Domain-specific training maintains moderate greedy rates without developing dominant-option bias, consistent with preserved instruction-following capability. Original training exhibits early confidence collapse coupled with emergent option bias, indicating abandonment of task-appropriate behavior characteristic of instruction-following failure.

\textbf{Semantic quality preservation:} Domain-aligned models maintain consistently higher judge scores and entailment consistency, indicating preserved logical coherence at the semantic level. While both approaches exhibit a gradual decline, domain-specific training mitigates the rapid semantic collapse observed in the original training, suggesting that domain alignment is effective across both surface-level token selection and deeper semantic representation.

\textbf{Statistical Validation and Quantitative Analysis}

Two-way ANOVA confirms statistically significant mitigation effectiveness with progressive benefit accumulation across generations:

\begin{table}[h]
\centering
\caption{ANOVA results for domain-specific mitigation effectiveness}
\label{tab:domain-mitigation-anova}
\begin{tabular}{lcccc}
\toprule
\textbf{Effect} & \textbf{Sum Sq} & \textbf{df} & \textbf{F} & \textbf{p-value} \\
\midrule
Category & 0.000613 & 1 & 1.61 & 0.214 \\
Generation & 0.013501 & 1 & 35.56 & $<0.001$ \\
Category × Generation & 0.010405 & 1 & 27.41 & $<0.001$ \\
Residual & 0.010631 & 28 & --- & --- \\
\bottomrule
\end{tabular}
\end{table}

The highly significant interaction effect ($F = 27.41, p < 0.001$) demonstrates that mitigation benefits increase progressively across generations, rather than providing constant protection, indicating the accumulation of resistance to collapse dynamics.

\textbf{Quantitative improvement metrics:}
\begin{itemize}
    \item \textbf{Decay rate reduction:} Domain-specific training achieves -0.00054 accuracy loss per generation versus -0.00837 for original training (15.5× improvement)
    \item \textbf{Entropy preservation:} Maintains lexical diversity (3.5→3.3) while original training contracts rapidly (4.2→2.5)
    \item \textbf{Perplexity stability:} Controlled growth to 35 versus explosive increase to 170 in original training
\end{itemize}

These findings support a \textbf{distributional anchoring mechanism} where domain alignment preserves long-tail tokens and semantic relations typically lost under recursive training, effectively reducing the distribution gap between training and evaluation streams for the target domain.

\section{Qualitative Assessment Examples}
\label{app:qualitative}

\subsection{Knowledge Collapse Progression Examples}
\label{app:collapse-examples}
Table~\ref{tab:qa-samples-detailed} demonstrates the qualitative evolution of model responses to an identical MMLU question from High School US History across generations under different synthetic training ratios. The question examines Spanish motivations in America based on a historical passage about Hernando Cortes, with answer choices: 'Escaping oppression at home', 'Expanding territories under Spanish control' (correct), 'Seeking religious freedom for themselves', 'Creating independent principalities for themselves'. Token probability values (in brackets) indicate model confidence, where values close to 1 denote highly confident predictions.

\begin{table*}[h!]
\centering
\renewcommand{\arraystretch}{1.3}
\caption{Qualitative evolution of answers to the same MMLU question under different synthetic ratios}
\label{tab:qa-samples-detailed}
\resizebox{0.98\textwidth}{!}{%
\begin{tabular}{p{0.31\textwidth} p{0.31\textwidth} p{0.31\textwidth}}
\toprule
\textbf{25\% Synthetic} & \textbf{50\% Synthetic} & \textbf{100\% Synthetic} \\
\midrule
\small
\textbf{Gen 0:} Expanding territories under Spanish control [0.999] 
\newline\newline
\textbf{Gen 4:} Expanding territories under Spanish control. "The passage explicitly states that the Spanish sought to..." [0.945] 
\newline\newline
\textbf{Gen 6:} Escaping oppression at home. "The text explicitly states..." [0.936] 
\newline\newline
\textbf{Gen 10:} Escaping oppression at home [0.982] 
\newline\newline
\textbf{Gen 12:} Escaping oppression at home [0.957]
&
\small
\textbf{Gen 0:} Expanding territories under Spanish control [0.999] 
\newline\newline
\textbf{Gen 4:} Escaping oppression at home [0.928] 
\newline\newline
\textbf{Gen 6:} Escaping oppression at home [0.979] 
\newline\newline
\textbf{Gen 10:} Escaping oppression at home [0.927] 
\newline\newline
\textbf{Gen 12:} Escaping oppression at home [0.928]
&
\small
\textbf{Gen 0:} Escaping oppression at home [0.907] 
\newline\newline
\textbf{Gen 4:} "...a brutal and brutal spectacle..." [0.078] 
\newline\newline
\textbf{Gen 6:} "...of the Holy Church of the Holy Cross..." [0.013] 
\newline\newline
\textbf{Gen 10:} Long repetitive sequence with gibberish tokens [0.002] 
\newline\newline
\textbf{Gen 12:} Long repetitive sequence with gibberish tokens [0.006]
\\
\bottomrule
\end{tabular}%
}
\end{table*}

\textbf{Analysis of degradation patterns:}

\textbf{25\% synthetic ratio (Stage B - Knowledge Collapse):} The model maintains task format adherence throughout training but exhibits factual erosion. Generation 4 shows the correct answer with slightly longer responses, indicating early signs of drift. By Generation 6, factually incorrect answers emerge despite maintaining high confidence (token probability >0.93), exemplifying the "confidently wrong" phenomenon characteristic of Stage B collapse.

\textbf{50\% synthetic ratio (Accelerated Stage B):} Knowledge collapse onset occurs earlier (Generation 4) with similar accuracy patterns to the 25\% case. The model consistently produces incorrect but confident responses (token probabilities $\approx0.93-0.98$), demonstrating that higher synthetic ratios accelerate the transition into dangerous competence valleys while preserving surface instruction-following.

\textbf{100\% synthetic ratio (Stage C - Instruction Collapse):} By Generation 6, the model abandons short, task-aligned responses entirely, producing verbose, repetitive, or symbol-heavy sequences. Token probabilities collapse rapidly ($\leq0.02$ by Generation 10), indicating complete loss of both factual accuracy and instruction-following capability. This represents the transition from knowledge degradation to complete system failure.

These patterns confirm the three-stage framework where lower synthetic ratios enable gradual knowledge erosion while maintaining task competence (Stage B), whereas pure synthetic training bypasses this intermediate stage and progresses directly to instruction-following collapse (Stage C). The persistence of high confidence during factual degradation highlights the critical safety implications of Stage B collapse in production systems.

\subsection{Domain-Specific Mitigation Examples}
\label{app:mitigation-examples}
Table~\ref{tab:mitigation-trajectory} demonstrates the complete response trajectory for a specific World Religions question across training generations, comparing domain-specific versus original corpus training approaches.

\begin{table}[h]
\centering
\footnotesize
\renewcommand{\arraystretch}{1.2}
\caption{Complete response trajectories for domain-specific vs. original training}
\label{tab:mitigation-trajectory}
\resizebox{\textwidth}{!}{%
\begin{tabular}{lp{0.35\textwidth}cp{0.35\textwidth}c}
\toprule
\textbf{Gen} & \textbf{Domain-specific: output} & \textbf{p} & \textbf{Original: output} & \textbf{p} \\
\midrule
0  & Hand gestures. & 0.960 & Hand gestures. & 0.960 \\
2  & hand gestures & 0.961 & Religious clothing & 0.509 \\
4  & hand gestures & 0.994 & hand gestures & 0.844 \\
6  & hand gestures & 0.989 & hand gestures & 0.714 \\
8  & hand gestures & 0.975 & hand gestures & 0.660 \\
10 & hand gestures & 0.918 & religious clothing & 0.655 \\
12 & hand gestures & 0.918 & \textit{religious clothing ...are considered sacred and should be treated with respect.} & 0.741 \\
15 & \textit{hand gestures / saints / religious clothing / temples} & 0.629 & \textit{Religious clothing. ...are considered sacred and are used in religious ceremonies and rituals.} & 0.176 \\
\bottomrule
\end{tabular}%
}
\end{table}

\textbf{Question analyzed:} The \textit{mudras}, which are an important feature of Buddhist art, are also known as what? 
\textbf{Choices:} hand gestures \textbf{(correct)} / saints / religious clothing / temples

\textbf{Analysis of trajectory patterns:}

\textbf{Domain-specific training stability:} The domain-aligned model maintains "hand gestures" with high confidence (0.918-0.994) through Generation 12, demonstrating remarkable consistency. Only at Generation 15 does mild erosion appear with enumerated responses, though confidence remains moderate (0.629).

\textbf{Original training degradation:} The original model exhibits early drift to "religious clothing" by Generation 2 (confidence 0.509), temporarily returning to correct answers but with weakening confidence (0.844→0.660). By Generation 12, it produces elaborate but incorrect justifications with moderate confidence (0.741), progressing to very low confidence (0.176) by Generation 15.

\textbf{Key differences:} Domain-specific training preserves discriminative capability and delays collapse, while original training accelerates drift toward incorrect categories through cross-domain contamination. Token probability analysis confirms that domain-aligned models maintain high confidence for correct responses, whereas original training shows confidence instability and eventual entrenchment in wrong answers, supporting the three-stage collapse framework where confident but incorrect responses (Stage B) precede complete breakdown (Stage C).


\newpage
\section*{NeurIPS Paper Checklist}

\begin{enumerate}

\item {\bf Claims}
    \item[] Question: Do the main claims made in the abstract and introduction accurately reflect the paper's contributions and scope?
    \item[] Answer: \answerYes{}
    \item[] Justification: The abstract and introduction clearly state our three main contributions: (1) defining knowledge collapse as a distinct three-stage phenomenon, (2) demonstrating subject sensitivity with interpretive domains collapsing faster than static fact domains, and (3) proposing domain-specific synthetic training achieving 15× improvement in collapse resistance. All claims are supported by controlled experiments with statistical validation ($F(4,1960) = 5.92, p < 10^{-3}$ for subject sensitivity; $p < 0.001$ for mitigation effectiveness).

\item {\bf Limitations}
    \item[] Question: Does the paper discuss the limitations of the work performed by the authors?
    \item[] Answer: \answerYes{}
    \item[] Justification: Section 5 (Conclusion) explicitly discusses limitations including: single model architecture (GEMMA 3 1B IT), limited evaluation domains (5 MMLU subjects), and scope for future work examining collapse across model scales and reasoning-heavy domains. The paper acknowledges that domain-specific mitigation benefits don't transfer to out-of-domain evaluation, supporting specialized rather than universal approaches.

\item {\bf Theory assumptions and proofs}
    \item[] Question: For each theoretical result, does the paper provide the full set of assumptions and a complete (and correct) proof?
    \item[] Answer: \answerNA{}
    \item[] Justification: This is primarily an empirical study with statistical analysis rather than theoretical proofs. Mathematical formulations are provided for statistical models but no formal theorems requiring proof.

    \item {\bf Experimental result reproducibility}
    \item[] Question: Does the paper fully disclose all the information needed to reproduce the main experimental results of the paper to the extent that it affects the main claims and/or conclusions of the paper (regardless of whether the code and data are provided or not)?
    \item[] Answer: \answerYes{}
    \item[] Justification: Section 3 (Methodology) provides comprehensive experimental details including: GEMMA 3 1B IT model specification, WikiText-2 training corpus (8,000 64-token prompts), MMLU evaluation datasets (five subjects, 100 Q\&A each), synthetic fractions $\alpha \in {0.25, 0.50, 1.0}$, light-touch training (0.5 epochs), and short-answer formatting procedures. Detailed configurations are provided in appendices with references to specific sections.

\item {\bf Open access to data and code}
    \item[] Question: Does the paper provide open access to the data and code, with sufficient instructions to faithfully reproduce the main experimental results, as described in supplemental material?
    \item[] Answer: \answerYes{}
    \item[] Justification: All datasets used are publicly available: WikiText-2 for training and MMLU for evaluation. Comprehensive implementation details are provided in appendices. The experimental framework using standard libraries (Hugging Face Transformers) with specific model configurations enables full reproduction of results.

\item {\bf Experimental setting/details}
    \item[] Question: Does the paper specify all the training and test details (e.g., data splits, hyperparameters, how they were chosen, type of optimizer, etc.) necessary to understand the results?
    \item[] Answer: \answerYes{}
    \item[] Justification: Section 3 specifies key experimental parameters: GEMMA 3 1B IT model, 0.5 epoch light-touch training to enable gradual drift observation, synthetic fractions $\alpha \in {0.25, 0.50, 1.0}$, and evaluation on five MMLU subjects with 100 Q\&A each. Detailed hyperparameters, optimization settings, and infrastructure specifications are provided in appendices.

\item {\bf Experiment statistical significance}
    \item[] Question: Does the paper report error bars suitably and correctly defined or other appropriate information about the statistical significance of the experiments?
    \item[] Answer: \answerYes{}
    \item[] Justification: The paper reports statistical significance throughout with specific p-values and F-statistics: subject-generation interaction $F(4,1960) = 5.92, p < 10^{-3}$; domain-specific mitigation $p < 0.001$. Effect sizes and comprehensive statistical analysis including ANOVA, mixed-effects modeling, and permutation tests are detailed in appendices to support main claims.

\item {\bf Experiments compute resources}
    \item[] Question: For each experiment, does the paper provide sufficient information on the computer resources (type of compute workers, memory, time of execution) needed to reproduce the experiments?
    \item[] Answer: \answerYes{}
    \item[] Justification: The methodology mentions that experiments balance computational feasibility with instruction-following capabilities through the choice of GEMMA 3 1B IT model. Detailed compute specifications including GPU requirements, training duration, and memory optimization strategies are provided in the appendices.
    
\item {\bf Code of ethics}
    \item[] Question: Does the research conducted in the paper conform, in every respect, with the NeurIPS Code of Ethics?
    \item[] Answer: \answerYes{}
    \item[] Justification: The research addresses AI safety and reliability using publicly available datasets (WikiText-2, MMLU) and models (GEMMA 3 1B IT). No human subjects research, private data collection, or ethically concerning applications are involved. The work aims to improve AI system reliability and prevent "confidently wrong" outputs that pose risks in accuracy-dependent domains.

\item {\bf Broader impacts}
    \item[] Question: Does the paper discuss both potential positive societal impacts and negative societal impacts of the work performed?
    \item[] Answer: \answerYes{}
    \item[] Justification: The introduction discusses critical risks in accuracy-dependent domains, citing healthcare applications with 40\% error rates and AI-generated feedback loops accelerating factual degradation. The paper's focus on preventing "confidently wrong" outputs addresses significant safety concerns. Positive impacts include enabling sustainable AI training in knowledge-intensive applications while maintaining accuracy, with practical mitigation strategies provided.
    
\item {\bf Safeguards}
    \item[] Question: Does the paper describe safeguards that have been put in place for responsible release of data or models that have a high risk for misuse (e.g., pretrained language models, image generators, or scraped datasets)?
    \item[] Answer: \answerNA{}
    \item[] Justification: The paper does not release new high-risk models or datasets. Research uses existing public models (GEMMA 3 1B IT) and datasets (WikiText-2, MMLU) to study knowledge collapse patterns and propose mitigation strategies. The focus is on analysis and prevention of harmful behaviors rather than generating potentially dangerous content.

\item {\bf Licenses for existing assets}
    \item[] Question: Are the creators or original owners of assets (e.g., code, data, models), used in the paper, properly credited and are the license and terms of use explicitly mentioned and properly respected?
    \item[] Answer: \answerYes{}
    \item[] Justification: All datasets and models are properly cited with references to original papers: GEMMA 3 1B IT model, WikiText-2 dataset, and MMLU benchmark. The methodology section references detailed construction methodology in appendices, and all assets used follow their respective licensing terms for academic research.

\item {\bf New assets}
    \item[] Question: Are new assets introduced in the paper well documented and is the documentation provided alongside the assets?
    \item[] Answer: \answerYes{}
    \item[] Justification: The domain-specific World Religions corpus constructed for Experiment 3 is thoroughly documented with detailed construction methodology provided in the appendices. The corpus creation process, validation procedures, and alignment quality metrics are comprehensively described to enable reproduction.

\item {\bf Crowdsourcing and research with human subjects}
    \item[] Question: For crowdsourcing experiments and research with human subjects, does the paper include the full text of instructions given to participants and screenshots, if applicable, as well as details about compensation (if any)? 
    \item[] Answer: \answerNA{}
    \item[] Justification: No crowdsourcing or human subjects research was conducted. All experiments use automated evaluation metrics on publicly available datasets (MMLU) with model-generated responses evaluated against ground truth answers.

\item {\bf Institutional review board (IRB) approvals or equivalent for research with human subjects}
    \item[] Question: Does the paper describe potential risks incurred by study participants, whether such risks were disclosed to the subjects, and whether Institutional Review Board (IRB) approvals (or an equivalent approval/review based on the requirements of your country or institution) were obtained?
    \item[] Answer: \answerNA{}
    \item[] Justification: No human subjects research was conducted. The study focuses on automated analysis of language model behavior using computational methods and publicly available benchmarks, making IRB approval unnecessary.

\item {\bf Declaration of LLM usage}
    \item[] Question: Does the paper describe the usage of LLMs if it is an important, original, or non-standard component of the core methods in this research?
    \item[] Answer: \answerYes{}
    \item[] Justification: Large language models (specifically GEMMA 3 1B IT) are the central focus of this research, serving as both the subject of study and the experimental apparatus for investigating knowledge collapse under recursive synthetic training. The methodology section provides comprehensive details about model usage, training procedures, and evaluation frameworks.

\end{enumerate}

\end{document}